\documentclass{article}
\usepackage{spconf,amsmath,epsfig}
\usepackage{multirow}
\usepackage{algorithmic}
\usepackage{algorithm}
\usepackage{float}
\usepackage{xcolor}
 \usepackage{graphicx} 
\let\OLDthebibliography\thebibliography
\renewcommand\thebibliography[1]{
  \OLDthebibliography{#1}
  \setlength{\parskip}{0pt}
  \setlength{\itemsep}{0pt plus 0.3ex}
}

\begin{document}\sloppy

\def\x{{\mathbf x}}
\def\L{{\cal L}}

\title{Improving the JPEG-resistance of Adversarial Attacks on Face Recognition by Interpolation Smoothing}
%
 \name{Kefu Guo, Fengfan Zhou, Hefei Ling$^{\dagger}$\thanks{$^\dagger$ Corresponding Author (lhefei@hust.edu.cn).}\thanks{This work was supported in part by the Natural Science Foundation of China under Grant 61972169,62372203 and 62302186, in part by the National key research and development program of China(2022YFB2601802), in part by the Major Scientific and Technological Project of Hubei Province (2022BAA046, 2022BAA042), in part by the Knowledge Innovation Program of Wuhan-Basic Research, in part by China Postdoctoral Science Foundation 2022M711251.}, Ping Li, Hui Liu}
 
 \address{School of Computer Science and Technology, Huazhong University of Science and Technology, China}
\maketitle

\begin{abstract}
JPEG compression can significantly impair the performance of adversarial face examples, which previous adversarial attacks on face recognition (FR) have not adequately addressed. Considering this challenge, we propose a novel adversarial attack on FR that aims to improve the resistance of adversarial examples against JPEG compression. Specifically, during the iterative process of generating adversarial face examples, we interpolate the adversarial face examples into a smaller size. Then we utilize these interpolated adversarial face examples to create the adversarial examples in the next iteration. Subsequently, we restore the adversarial face examples to their original size by interpolating. Throughout the entire process, our proposed method can smooth the adversarial perturbations, effectively mitigating the presence of high-frequency signals in the crafted adversarial face examples that are typically eliminated by JPEG compression. Our experimental results demonstrate the effectiveness of our proposed method in improving the JPEG-resistance of adversarial face examples.
\end{abstract}
\section{Introduction}
\label{sec:intro}

FR has been widely used in various areas such as security and payment systems. However, many works~\cite{Komkov_2021, hu2022protecting} has demonstrated that FR models can be deceived by adding adversarial perturbations on the input images. These modified images, known as adversarial face examples, pose a significant threat to the security and reliability of existing FR systems. Hence, to ensure the safer deployment of FR, it becomes important to enhance the performance of adversarial face examples, thereby revealing the vulnerabilities and blind spots present in current FR models.

The adversarial examples generated by the existing adversarial attacks~\cite{goodfellow2015explaining, BIM, MI} on FR often contain a significant amount of high-frequency signals. However, digital images will be JPEG compressed in many channels to facilitate faster storage and transmission. Previous works~\cite{dziugaite2016study, guo2018countering} have shown that JPEG compression discards high-frequency signals in the image and introduces distortion that significantly affects the adversarial perturbations. As a result, the attack success rate (ASR) of adversarial examples significantly decreases. Futhermore, in certain FR systems that employ defense strategies, compression-based defense methods (e.g., JPEG~\cite{dziugaite2016study}, FD~\cite{FD}, Comdefend~\cite{jia2019comdefend}, etc.) may be used to preprocess digital images to reduce the vulnerability of FR systems to adversarial attacks. 

\begin{figure}[tb]
\centerline{\includegraphics[width=8cm]{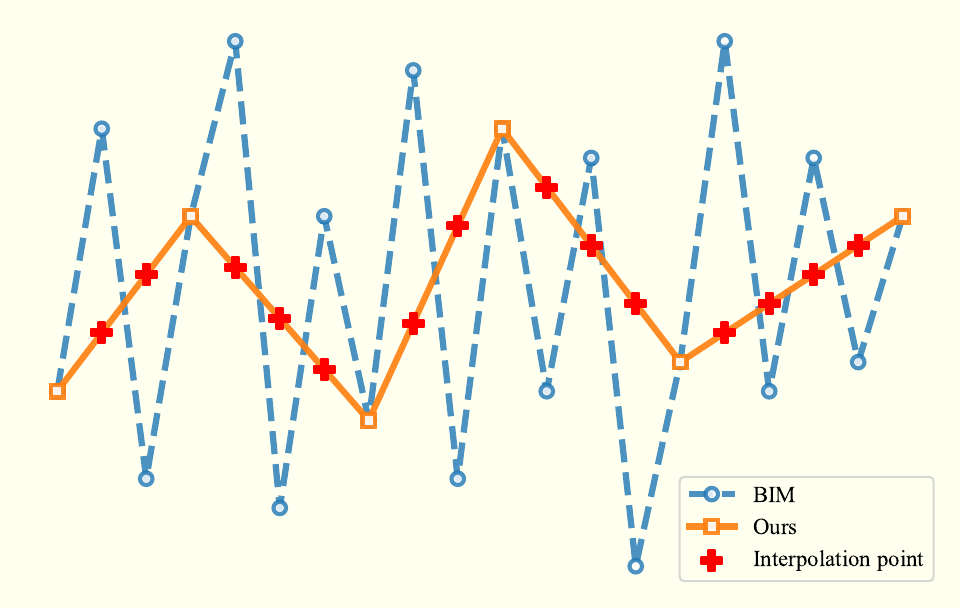}}
\caption{Comparison of pixel values between the adversarial examples generated by BIM and our method. The horizontal axis represents the neighboring pixel points of adversarial examples, and the vertical axis represents the corresponding  pixel values.}
\label{fig1}
\end{figure}

Currently, there is limited research focused on generating adversarial face examples that can resist JPEG compression. In this paper, we propose the Interpolation Attack Method (IAM) to improve the JPEG-resistance of adversarial attacks on FR. Our proposed IAM applies interpolation to craft smoother images, resulting in less high-frequency signals compared with traditional gradient-based methods, as shown in Fig.\ref {fig1}. The main contributions of our work are summarized as follows:

1. We explore the influence of different JPEG compression levels on the ASR of adversarial attacks applied to multiple FR models.

2. We propose IAM to improve the JPEG-resistance of adversarial attacks, and our approach is compatible with the majority of existing adversarial attack methods.

3. We combine IAM with multiple existing attack methods and conduct comprehensive experiments on multiple FR models. The results demonstrate that our proposed approach effectively improves the JPEG-resistance of adversarial attacks.

\section{Related Works}
\subsection{Adversarial Attacks}

Our proposed IAM primarily focuses on improving the JPEG-resistance of gradient-based attacks.
Most gradient-based adversarial attacks are derived from the FGSM~\cite{goodfellow2015explaining}  and have made subsequent improvements.
BIM~\cite{BIM} is essentially a iterative version of FGSM, but it is prone to poor local optimization and overfitting. MI~\cite{MI} introduced momentum to escape from the poor local optimization and NI-FGSM~\cite{lin2020nesterov} uses Nesterov to accelarate it. 

To improve the transferability of adversarial examples, DI~\cite{DI} introduces input diversity. DFANet~\cite{DFANet} incorporates dropout layers into the surrogate attacker model during the generation process of adversarial examples and SSA~\cite{ssa} applies transformations to the input image by introducing Gaussian noise and random masks in the frequency domain.
\subsection{JPEG-resistance of Adversarial Samples}
\label{ssec:subhead}
Most of the aforementioned adversarial attacks lack the JPEG-resistance, making them ineffective on FR models when adversarial samples undergo compression using methods like JPEG compression~\cite{dziugaite2016study, guo2018countering}.
Shin and Song~\cite{JPEGSS} proposed an adaptive attack method (we denote it as JPEGSS in the following) with a differentiable JPEG approximation. The adversarial examples generated by JPEGSS are able to resist JPEG compression to some extent. However, this work require the attacker obtain the quality factor (QF) of the JPEG compression that attacker cannot obtain in practical application. Therefore, we focus on improving the JPEG-resistance without obtaining QF.

\begin{figure}[htb]
\centerline{\includegraphics[width=8cm]{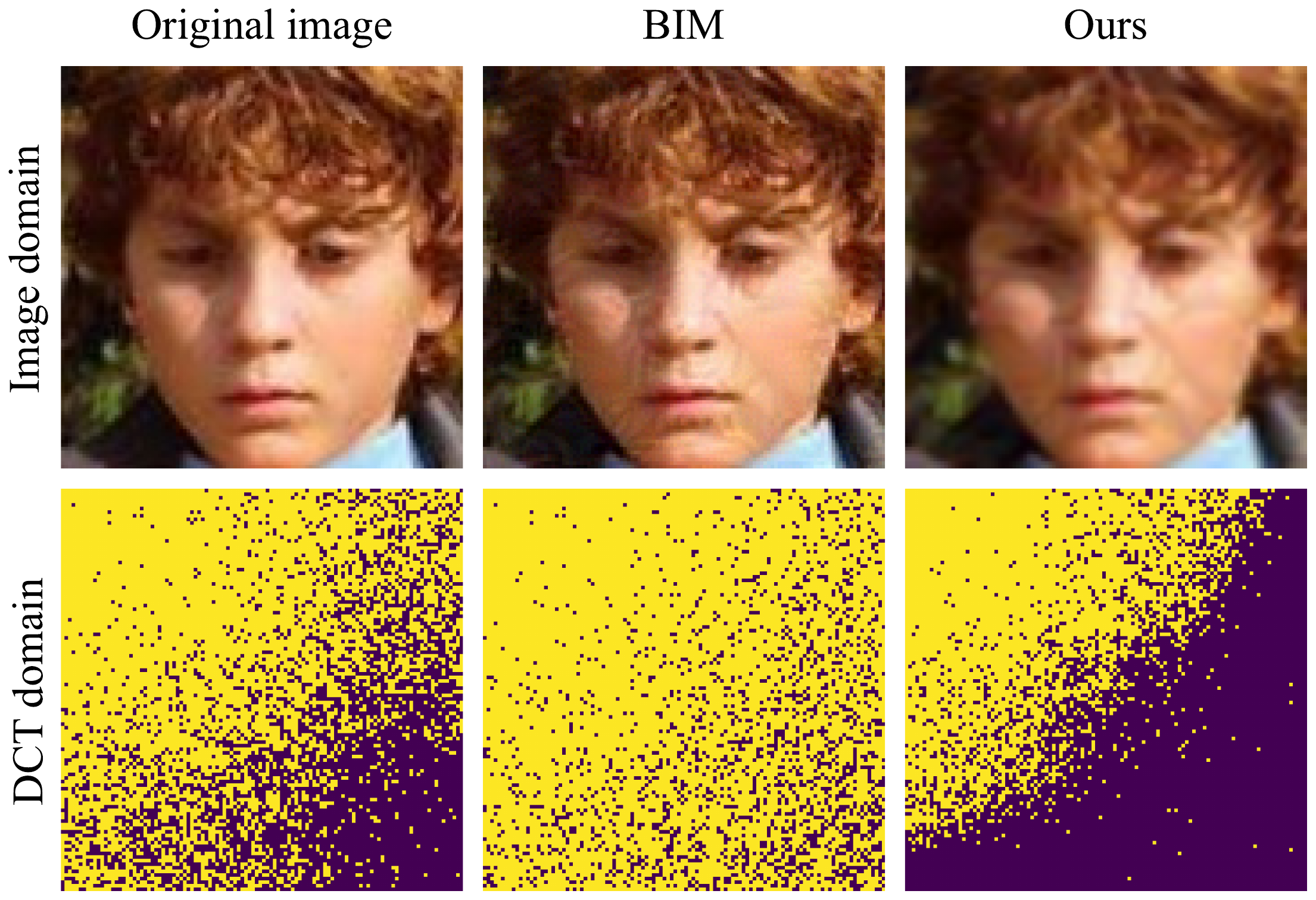}}
\caption{Images and their corresponding DCT coefficients. The images in the first row represent the original image, adversarial examples generated by BIM and our method. The second row are their corresponding DCT coefficients. In the DCT domain, the upper left corner is the location of the lowest frequency, and the frequency gradually increases in other locations. Light color indicates larger coefficient and dark color indicates smaller coefficient.}
\label{fig2}
\end{figure}

The JPEG compression process discards some of the high-frequency signals in Discrete Cosine Transform (DCT) domain~\cite{guo2018countering}. We perform DCT on the entire image. As shown in Fig.\ref{fig2}, compared to original image, BIM introduces additional high-frequency signals to the image which are more likely to be discarded during JPEG compression. Consequently, BIM exhibits vulnerability to JPEG compression due to the potential loss of these high-frequency signals. On the contrary, there are less high-frequency signals in the adversarial example generated by our proposed approach so that there will be less information loss during JPEG compression.

\section{Methodology}
Let $\mathcal{F}(x)$ denote an FR model to extract the embedding vector from a face image $x$, $x^{s}$ denotes the image of the attacker and $x^{t}$ denotes the victim image of the victim. The optimization objective of the attacks in our paper is to deceive $\mathcal{F}$ to recognize JPEG compressed $x^{adv}$ as $x^{t}$ while ensuring $x^{adv}$ visually resembles $x^{s}$. The objective can be expressed as:
\begin{equation}
  \begin{aligned}
    x^{adv}=&\mathop{\arg\min}\limits_{x^{adv}}(\mathcal{D}(\mathcal{F}(\mathcal{J}\left(x^{adv}\right)), \mathcal{F}(x^{t}))),\\
    &s.t. \left \| x^{adv}-x^{s} \right \| _{p}\le \epsilon
    \label{eq:opt_obj}
  \end{aligned}
\end{equation}
\noindent where $\mathcal{D}$ is a pre-selected distance metric, $\mathcal{J}$ is a JPEG compression function, and $\epsilon$ is the maximum perturbation magnitude.

A JPEG image is formed by compression of the quantized DCT coefficients~\cite{guo2018countering}, and some high-frequency signals will be discarded during compression.
As shown in Fig.\ref{fig2}, there are more larger coefficients in high-frequency region in the DCT domain of the adversarial example generated by BIM~\cite{BIM}. 
To craft adversarial face example that are resistant to the JPEG compression, it is important to reduce the high-frequency signals in the adversarial face examples.
In our paper, we use interpolation to smooth the adversarial face examples to reduce the the high-frequency signals in the crafted adversarial face examples.
Our proposed method can be combined with most gradient-based adversarial attacks. To make our illustration more clear. We only introduce our proposed method with BIM.
Specifically, the loss function can be expressed as:
\begin{equation}
  \mathcal{L}\left(x^{adv}, x^t\right)=\Vert\phi\left(\mathcal{F}\left(x^{adv}\right)\right)-\phi\left(\mathcal{F}\left(x^{t}\right)\right)\Vert^2_2\label{eq:bpfa_loss},
\end{equation}
where $\phi(x)$ denotes the operation that normalizes $x$. $x^{adv}$ is the adversarial example that initiates with the same value of $x^{s}$.

Then we use the bilinear interpolation to smooth the adversarial face examples.
Bilinear interpolation is a linear interpolation extension of an interpolation function with two variables, and its core idea is to perform a linear interpolation in each direction. Bilinear interpolation calculates the pixel value of 1 point in the interpolated image from the four neighboring points in the original image. Let the pixel value of any point $(\alpha,\beta)$ in the interpolated image be $f(\alpha,\beta)$. Suppose the coordinates of the neighboring points are $(\alpha_{1}, \beta_{1})$, $(\alpha_{1}, \beta_{2})$, $(\alpha_{2}, \beta_{1})$ and $(\alpha_{2}, \beta_{2})$. First, linear interpolate in the $x$ direction:
\begin{equation}
  \begin{aligned}
    f(\alpha, \beta_{1})=\frac{\alpha_{2}-\alpha}{\alpha_{2}-\alpha_{1}}f(\alpha_{1}, \beta_{1})+\frac{\alpha-\alpha_{1}}{\alpha_{2}-\alpha_{1}}f(\alpha_{2}, \beta_{1}),\\
    f(\alpha, \beta_{2})=\frac{\alpha_{2}-\alpha}{\alpha_{2}-\alpha_{1}}f(\alpha_{1}, \beta_{2})+\frac{\alpha-\alpha_{1}}{\alpha_{2}-\alpha_{1}}f(\alpha_{2}, \beta_{2}),
  \end{aligned}
\end{equation}
\noindent where $f(\alpha_{1}, \beta_{1})$, $f(\alpha_{1}, \beta_{2})$, $f(\alpha_{2}, \beta_{1})$ and $f(\alpha_{2}, \beta_{2})$ are the pixel values of the four
neighboring points. Then, linear interpolate in the $y$ direction:
\begin{equation}
  \begin{aligned}
    f(\alpha,\beta)= \frac{\beta_{2}-\beta}{\beta_{2}-\beta_{1}}f(\alpha, \beta_{1})+\frac{\beta-\beta_{1}}{\beta_{2}-\beta_{1}}f(\alpha, \beta_{2}).
  \end{aligned}
\end{equation}

Then, we define the distance between two neighboring points after interpolation as the interpolation factor $ f_{inter} $. When the $ f_{inter} $ is fixed:
\begin{equation}
  \begin{aligned}
    \alpha_{2}-\alpha_{1}=\beta_{2}-\beta_{1}, f_{inter}:=\alpha_{2}-\alpha_{1}.
    \nonumber
  \end{aligned}
\end{equation}

Let $\mathcal{I}(\cdot, \cdot)$ denotes bilinear interpolation, and interpolated image be $\tilde{x}$. We can express the interpolated image as $\tilde{x} = \mathcal{I}(x, f_{inter})$, and for a point $(\alpha,\beta)$ in $\tilde{x}$: 
\begin{equation}
  \begin{aligned}
        &F=\begin{bmatrix} f(\alpha_{1}, \beta_{1}) & f(\alpha_{1}, \beta_{2})\\f(\alpha_{2}, \beta_{1}) & f(\alpha_{2}, \beta_{2})\end{bmatrix},\\
    f(\alpha,\beta)=&\frac{1}{f_{inter}^{2}} \cdot 
    \begin{bmatrix}\alpha_{2}-\alpha  & \alpha-\alpha_{1}\end{bmatrix} \cdot F \cdot
    \begin{bmatrix}
      \beta_{2}-\beta \\
      \beta-\beta_{1}
    \end{bmatrix}.
  \end{aligned}
\end{equation}

Let $x_{t}^{adv}$ be an adversarial face example in the $t$-th iteration.
To craft the adversarial face examples with high JPEG-resistance, we use $\mathcal{I}$ to downsample the adversarial face example $x^{adv}_{t-1}$ by a ratio $f_{inter}$ that is lower than 1:
\begin{equation}
  \tilde{x}^{adv}_{t-1} = \mathcal{I}\left(x^{adv}_{t-1}, f_{inter}\right).
 \label{downsample}
\end{equation}

After getting $\tilde{x}^{adv}_{t-1}$, we can calculate the adversarial perturbations $\tilde{p}^{adv}_{t-1}$ on $\tilde{x}^{adv}_{t-1}$ by backpropagation:
\begin{equation}
  \tilde{p}^{adv}_{t-1}={\rm sign}\left(\nabla_{\tilde{x}^{adv}_{t-1}}\mathcal{L}\left(\tilde{x}^{adv}_{t-1}, x^t\right)\right).
\end{equation}

The temporary adversarial face example $\tilde{x}^{adv}_{t}$ can be calculate using the following formula:
\begin{equation}
  \tilde{x}^{adv}_{t}=\tilde{x}^{adv}_{t-1}-\beta \tilde{p}^{adv}_{t-1},
\end{equation}
where $\beta$ is the stepsize for crafting the adversarial face examples.

To improve the smoothness of crafted adversarial face example, we interpolate $\tilde{x}^{adv}_{t}$ by ratio $\frac{1}{f_{inter}}$:
\begin{equation}
{x}^{adv}_{t}=\mathcal{I}\left(\tilde{x}^{adv}_{t},\frac{1}{f_{inter}}\right),
\label{interpolation}
\end{equation}
${x}^{adv}_{t}$ is smoother than the original adversarial example of BIM, therefore more adversarial perturbations will be remained after being JPEG compressed. We summarize the algorithm of IAM integrated into BIM for Impersonation Attacks in Algorithm \ref{alg:iam}. 

\begin{algorithm}[tb]
  \renewcommand{\algorithmicrequire}{\textbf{Input:}}
  \renewcommand{\algorithmicensure}{\textbf{Output:}}
  \caption{Interpolation Attack Method} 
  \label{alg:iam} 
  \begin{algorithmic}[1]
    \REQUIRE Negative face image pair $\{x^s, x^t\}$, the interpolation factor $f_{inter}$ ($f_{inter}<1$), the maximum number of iterations $N_{max}$, the stepsize for crafting the adversarial face examples $\beta$, the surrogate FR model $\mathcal{F}$.
    \ENSURE An adversarial face example $x^{adv}$
    \STATE $x^{adv}_0=x^{s}$
    \FOR{$t=1 ,..., N_{max}$}
    \STATE \textcolor{blue}{\# Downsample the adversarial face example by Eq.\ref{downsample}}\\ $\tilde{x}^{adv}_{t-1} = \mathcal{I}\left(x^{adv}_{t-1}, f_{inter}\right)$
    \STATE $\tilde{p}^{adv}_{t-1}={\rm sign}\left(\nabla_{\tilde{x}^{adv}_{t-1}}\mathcal{L}\left(\tilde{x}^{adv}_{t-1}, x^t\right)\right)$
    \STATE $\tilde{x}^{adv}_{t}=\tilde{x}^{adv}_{t-1}-\beta \tilde{p}^{adv}_{t-1}$ 
    \STATE \textcolor{blue}{\# Interpolate the adversarial face example by Eq.\ref{interpolation}}\\
    $x^{adv}_{t} = \mathcal{I}(\tilde{x}^{adv}_{t}, \frac{1}{f_{inter}})$
    \ENDFOR
    \RETURN $x^{adv}=x^{adv}_{N_{max}}$
  \end{algorithmic} 
\end{algorithm}

\begin{table*}[tb]
  \begin{center}
  \caption{ASRs of impersonation attacks on CelebA-HQ with the state-of-the-art attacks as the baseline. The first column reprents the different JPEG compression extends. The third to fifth columns in the first row represent the normally trained victim models, and the sixth to eighth columns in the first row represent the adversarially trained victim models.}
        \label{table1}
      \begin{tabular}{c|c|c|c|c|c|c|c} 
          \hline
            JPEG Images & Attack & IR152 & MF & FaceNet & IR152$_{adv}$ & MF$_{adv}$ & FaceNet$_{adv}$\\
          \hline
          \multirow{5}{*}{Uncompressed} 
          & BIM / +IAM & 40.3 / \textbf{56.3} & 84.1 / \textbf{88.7} & 15.4 / \textbf{23.2} & 21.2 / \textbf{28.0} & 8.7 / \textbf{15.5} & 4.2 / \textbf{7.0}\\
          & MI / +IAM & 33.2 / \textbf{46.4} & 73.6 / \textbf{82.3} & 16.3 / \textbf{21.0} & 10.8 / \textbf{16.1} & 10.5 / \textbf{13.6} & 5.6 / \textbf{7.5}\\
          & DI / +IAM & 24.0 / \textbf{63.0} & 92.1 / \textbf{97.9} & 31.9 / \textbf{45.6} & 10.3 / \textbf{28.8} & 36.0 / \textbf{37.7} & 8.0 / \textbf{13.5}\\
          & DFANet / +IAM & 39.1 / \textbf{56.1} & 92.7 / \textbf{95.1} & 21.3 / \textbf{31.6} & 11.5 / \textbf{20.9} & 12.1 / \textbf{20.8} & 5.4 / \textbf{7.9}\\
          & SSA / +IAM & 57.9 / \textbf{64.6} & \textbf{97.3} / \textbf{97.3} & 34.1 / \textbf{38.6} & 19.3 / \textbf{24.2} & 23.1 / \textbf{30.0} & 8.7 / \textbf{11.3}\\
          \hline
          \multirow{5}{*}{QF $=75$} 
          & BIM / +IAM & 35.3 / \textbf{54.4} & 77.6 / \textbf{86.6} & 13.6 / \textbf{21.0} & 8.5 / \textbf{15.9} & 7.7 / \textbf{15.1} & 4.2 / \textbf{6.4}\\
          & MI / +IAM & 26.0 / \textbf{41.6} & 55.5 / \textbf{73.6} & 11.9 / \textbf{17.3} & 8.1 / \textbf{14.1} & 6.9 / \textbf{11.1} & 3.7 / \textbf{6.7}\\
          & DI / +IAM & 19.8 / \textbf{61.2} & 95.6 / \textbf{98.7} & 25.0 / \textbf{41.1} & 8.2 / \textbf{26.5} & 28.0 / \textbf{32.1} & 5.8 / \textbf{12.2}\\
          & DFANet / +IAM & 30.4 / \textbf{51.0} & 79.6 / \textbf{92.2} & 15.4 / \textbf{28.2} & 8.6 / \textbf{18.5} & 8.2 / \textbf{17.6} & 4.0 / \textbf{7.0}\\
          & SSA / +IAM & 50.4 / \textbf{62.5} & 89.5 / \textbf{95.3} & 27.0 / \textbf{35.0} & 14.7 / \textbf{22.3} & 16.7 / \textbf{25.2} & 6.9 / \textbf{10.0}\\
          \hline
          \multirow{5}{*}{QF $=50$} 
          & BIM / +IAM & 31.1 / \textbf{49.0} & 73.6 / \textbf{81.5} & 12.7 / \textbf{18.3} & 8.3 / \textbf{15.3} & 7.1 / \textbf{13.3} & 3.9 / \textbf{6.2}\\
          & MI / +IAM & 32.2 / \textbf{36.6} & 51.2 / \textbf{67.1} & 11.6 / \textbf{15.3} & 7.1 / \textbf{12.0} & 6.6 / \textbf{9.7} & 3.7 / \textbf{6.1}\\
          & DI / +IAM & 19.2 / \textbf{59.6} & 94.0 / \textbf{98.3} & 24.0 / \textbf{39.9} & 8.3 / \textbf{25.7} & 25.6 / \textbf{29.9} & 5.3 / \textbf{12.1}\\
          & DFANet / +IAM & 26.4 / \textbf{49.1} & 75.1 / \textbf{89.9} & 15.1 / \textbf{23.2} & 7.7 / \textbf{16.5} & 7.3 / \textbf{16.4} & 4.0 / \textbf{6.0}\\
          & SSA / +IAM & 49.1 / \textbf{60.2} & 87.7 / \textbf{93.2} & 25.5 / \textbf{31.5} & 15.0 / \textbf{21.0} & 15.9 / \textbf{23.0} & 6.5 / \textbf{8.9}\\
          \hline
                \multirow{5}{*}{QF $=25$} 
          & BIM / +IAM & 22.9 / \textbf{33.9} & 58.4 / \textbf{69.2} & 9.7 / \textbf{15.1} & 5.5 / \textbf{11.4} & 5.8 / \textbf{10.7} & 3.3 / \textbf{5.1}\\
          & MI / +IAM & 17.7 / \textbf{30.0} & 39.6 / \textbf{60.1} & 9.0 / \textbf{13.9} & 5.9 / \textbf{10.8} & 5.9 / \textbf{9.5} & 2.6 / \textbf{6.0}\\
          & DI / +IAM & 9.7 / \textbf{52.2} & 77.8 / \textbf{88.1} & 4.2 / \textbf{32.0} & 2.9 / \textbf{20.7} & 3.1 / \textbf{29.9} & 1.2 / \textbf{10.3}\\
          & DFANet / +IAM & 21.0 / \textbf{36.1} & 60.9 / \textbf{75.7} & 10.3 / \textbf{17.6} & 6.8 / \textbf{12.4} & 5.7 / \textbf{11.6} & 3.1 / \textbf{5.7}\\
          & SSA / +IAM & 41.3 / \textbf{51.3} & 80.1 / \textbf{86.0} & 21.0 / \textbf{26.5} & 12.4 / \textbf{17.5} & 12.6 / \textbf{18.5} & 5.3 / \textbf{8.2}\\
          \hline
        \end{tabular}
    \end{center}
\end{table*}
\begin{table*}[tb]
  \begin{center}
      \caption{ASRs of impersonation attacks on LFW with the state-of-the-art attacks as the baseline. The first column reprents the different JPEG compression extends. The third to fifth columns in the first row represent the normally trained victim models, and the sixth to eighth columns in the first row represent the adversarially trained victim models.}
    \label{table2}
      \begin{tabular}{c|c|c|c|c|c|c|c} 
          \hline
            JPEG Images & Attack & IR152 & MF & FaceNet & IR152$_{adv}$ & MF$_{adv}$ & FaceNet$_{adv}$\\
          \hline
          \multirow{5}{*}{Uncompressed} 
          & BIM / +IAM & 34.7 / \textbf{45.2} & 80.5 / \textbf{83.2} & 13.9 / \textbf{19.2} & 15.2 / \textbf{21.2} & 4.4 / \textbf{10.2} & 4.3 / \textbf{5.8}\\
          & MI / +IAM & 30.3 / \textbf{37.7} & 73.2 / \textbf{78.4} & 16.3 / \textbf{18.6} & 10.6 / \textbf{14.1} & 7.5 / \textbf{10.1} & 6.2 / \textbf{7.5}\\
          & DI / +IAM & 57.5 / \textbf{59.7} & 96.1 / \textbf{97.7} & 45.6 / \textbf{47.3} & 26.1 / \textbf{29.9} & 23.0 / \textbf{33.1} & 15.2 / \textbf{17.1}\\
          & DFANet / +IAM & 37.5 / \textbf{50.5} & 93.5 / \textbf{94.6} & 20.9 / \textbf{30.8} & 12.6 / \textbf{19.0} & 10.2 / \textbf{17.9} & 7.5 / \textbf{10.2}\\
          & SSA / +IAM & 59.1 / \textbf{61.4} & 97.0 / \textbf{97.2} & 38.7 / \textbf{39.9} & 23.3 / \textbf{25.3} & 19.5 / \textbf{26.6} & 12.8 / \textbf{13.8}\\
          \hline
          \multirow{5}{*}{QF $=75$} 
          & BIM / +IAM & 28.5 / \textbf{42.6} & 68.5 / \textbf{78.2} & 12.1 / \textbf{17.4} & 6.5 / \textbf{12.0} & 3.4 / \textbf{9.3} & 14.1 / \textbf{15.3}\\
          & MI / +IAM & 26.0 / \textbf{35.5} & 63.0 / \textbf{71.9} & 13.8 / \textbf{17.2} & 10.2 / \textbf{12.1} & 6.3 / \textbf{9.2} & 5.3 / \textbf{7.0}\\
          & DI / +IAM & 52.7 / \textbf{57.6} & 94.6 / \textbf{94.9} & 40.7 / \textbf{43.4} & 23.4 / \textbf{29.1} & 20.0 / \textbf{30.6} & 13.4 / \textbf{16.1}\\
          & DFANet / +IAM & 32.3 / \textbf{46.7} & 84.8 / \textbf{91.8} & 17.5 / \textbf{27.3} & 10.5 / \textbf{17.6} & 8.5 / \textbf{16.8} & 5.9 / \textbf{9.5}\\
          & SSA / +IAM & 52.2 / \textbf{58.8} & 93.3 / \textbf{93.8} & 34.3 / \textbf{36.7} & 20.0 / \textbf{23.8} & 17.6 / \textbf{24.8} & 11.5 / \textbf{12.7}\\
          \hline
          \multirow{5}{*}{QF $=50$} 
          & BIM / +IAM & 24.6 / \textbf{38.7} & 59.5 / \textbf{71.3} & 11.0 / \textbf{15.7} & 5.1 / \textbf{10.6} & 2.6 / \textbf{8.5} & 3.7 / \textbf{4.9}\\
          & MI / +IAM & 23.7 / \textbf{32.4} & 57.6 / \textbf{66.3} & 12.7 / \textbf{15.0} & 8.5 / \textbf{11.8} & 5.9 / \textbf{8.8} & 5.5 / \textbf{7.1}\\
          & DI / +IAM & 46.7 / \textbf{54.8} & 92.4 / \textbf{93.0} & 36.4 / \textbf{42.2} & 21.1 / \textbf{27.7} & 17.9 / \textbf{28.6} & 12.2 / \textbf{14.8}\\
          & DFANet / +IAM & 25.8 / \textbf{42.1} & 79.9 / \textbf{88.8} & 14.5 / \textbf{23.9} & 9.1 / \textbf{15.9} & 7.1 / \textbf{15.0} & 5.4 / \textbf{9.0}\\
          & SSA / +IAM & 45.8 / \textbf{55.4} & 90.8 / \textbf{92.1} & 30.5 / \textbf{34.2} & 17.5 / \textbf{21.0} & 15.5 / \textbf{22.4} & 10.3 / \textbf{11.0}\\
          \hline
                \multirow{5}{*}{QF $=25$} 
          & BIM / +IAM & 16.4 / \textbf{29.2} & 39.5 / \textbf{56.1} & 7.9 / \textbf{11.2} & 3.5 / \textbf{8.3} & 2.1 / \textbf{5.6} & 2.5 / \textbf{4.6}\\
          & MI / +IAM & 18.4 / \textbf{24.2} & 42.1 / \textbf{50.8} & 10.4 / \textbf{13.1} & 6.6 / \textbf{8.6} & 4.8 / \textbf{6.8} & 4.6 / \textbf{6.0}\\
          & DI / +IAM & 36.4 / \textbf{47.0} & 84.3 / \textbf{87.9} & 27.0 / \textbf{35.2} & 5.5 / \textbf{23.3} & 14.0 / \textbf{24.4} & 9.6 / \textbf{12.3}\\
          & DFANet / +IAM & 18.3 / \textbf{31.9} & 63.3 / \textbf{77.0} & 11.1 / \textbf{18.1} & 6.2 / \textbf{12.5} & 5.4 / \textbf{11.5} & 4.1 / \textbf{7.0}\\
          & SSA / +IAM & 36.2 / \textbf{46.1} & 82.4 / \textbf{85.6} & 22.8 / \textbf{28.0} & 12.2 / \textbf{17.1} & 12.0 / \textbf{19.0} & 8.3 / \textbf{10.1}\\
          \hline
        \end{tabular}
  \end{center}
\end{table*}
\section{Experiments}
\subsection{Experimental Setup}
\textbf{Datasets.} In our experiments, we chose CelebA-HQ~\cite{celebahq} and LFW~\cite{lfw} as the datasets. These two datasets are commonly used in the research of adversarial attacks on FR~\cite{hu2022protecting, xiao2021improving, bpfa}.
We randomly selected 1000 face pairs from the CelebA-HQ dataset and the LFW dataset respectively. All the pairs are negative pairs, where one image in a pair is used as the attacker image and the other as the victim image.

\textbf{FR Models.} We use three normally trained FR models including IRSE50, FaceNet, MobileFace (we denote it as MF in the following) and IR152 which are widely used in various applications and their adversarially trained version~\cite{mter}, namely FaceNet$_{adv}$, MF$_{adv}$ and IR152$_{adv}$. We choose the thresholds when FAR@0.001 on the CelebA-HQ dataset and the LFW dataset as the thresholds to calculate the attack success rate (ASR) for the impersonation attacks. 

\textbf{Baselines.} The state-of-the-art attack method on FR is the combination of BIM~\cite{BIM}, MI~\cite{MI}, DI~\cite{DI}, DFANet~\cite{DFANet} and SSA~\cite{ssa}. We selected these methods as our baselines for generating uncompressed adversarial examples.

\textbf{JPEG Compression.} 
To evaluate the JPEG-resistance of adversarial samples on FR models, we employed three different QF of 25, 50 and 75 during JPEG compression to obtain the corresponding JPEG images. To explore the influence of JPEG compression on adversarial samples, we conducted experiments using a range of QF values from 10 to 90.
\subsection{Comparison Study}
To evaluate the effectiveness of our proposed IAM, we conducted impersonation attack experiments on CelebA-HQ and LFW using state-of-the-art attacks. We set $f_{inter}=\frac{1}{2}$, $N_{max}=10$, $\beta=1.0$ and $\epsilon=10$. We utilized IRSE50 model as the attacker model to generate adversarial examples, while employing other FR models as victim models to conduct black-box attack experiments. All the generated adversarial examples undergo JPEG compression with QF of 25, 50 and 75, respectively. The results on CelebA-HQ and and LFW are demonstrated in Table.\ref{table1} and Table.\ref{table2}. In these two tables, the number before slash is the ASR of the baseline method, and the number after slash is the ASR of the baseline method with the combination of IAM. 

The results indicate that the attacks exhibit superior performance on FR models and their adversarially trained version when supplemented with IAM, compared with the attacks without the combination of IAM. Besides, the JPEG-resistance of IAM is manifest across three levels of image QF: high (75), medium (50), and low (25).
The results demonstrate that IAM can be well-combined with the existing state-of-the-art adversarial attacks to improve the JPEG-resistance of them on different datasets.
The reason is apparent: IAM effectively reduces the low-frequency signal in the crafted adversarial face examples, leading to a more JPEG-resistant outcome.

 Addtionally, it is noteworthy that the attacks supplemented with IAM remain better performance when the adversarial examples are uncompressed. This phenomenon suggests that IAM can be effectively integrated with the existing adversarial
attacks to enhance the transferability of them. The interpolation applied to images has a data-enrichment effect, leading to an enhancement in transferability.
\subsection{Ablation Study}
\begin{figure}[!t]
\centerline{\includegraphics[width=8.65cm]{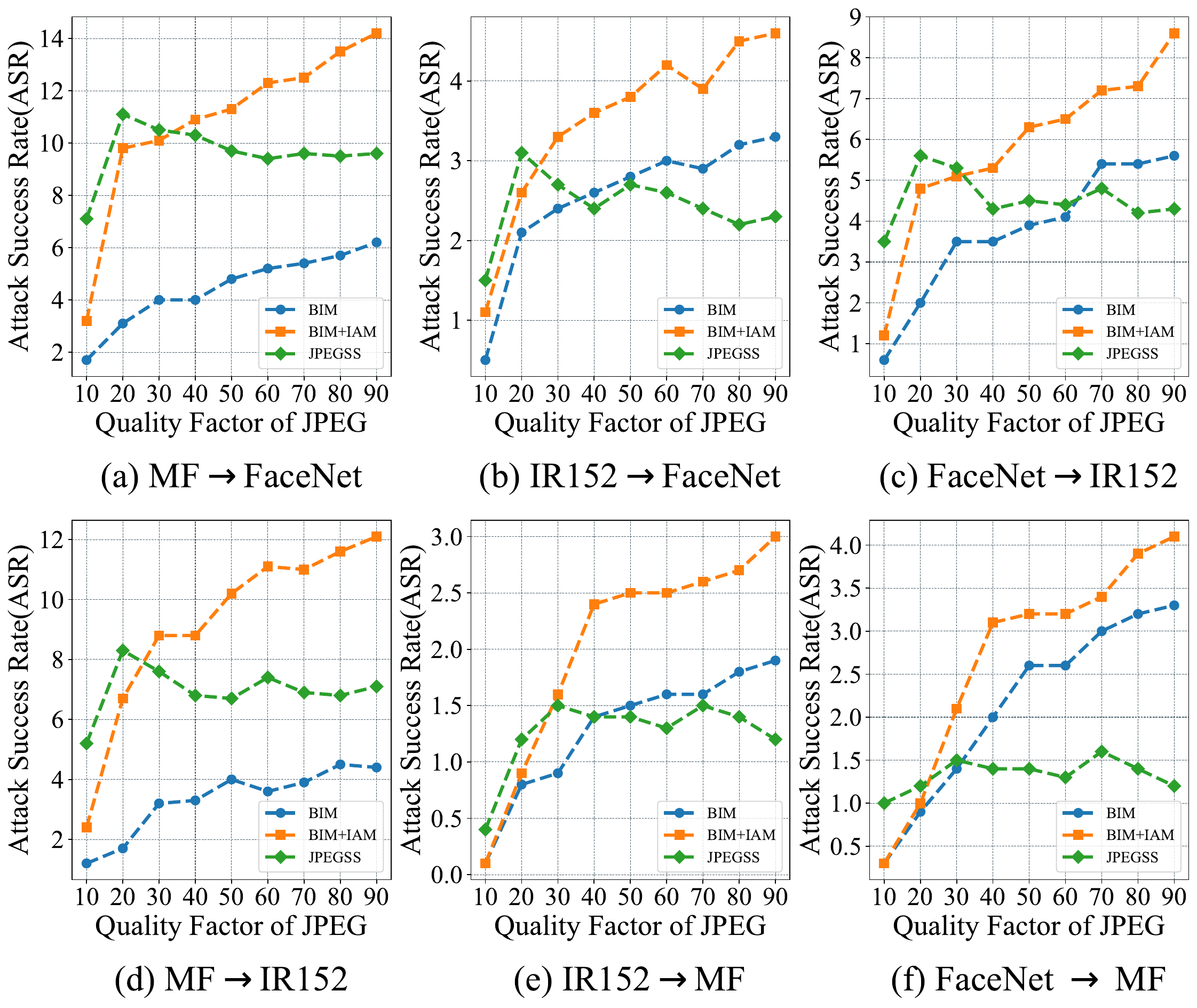}}
\caption{ASRs of the JPEG adversarial examples with different QF. The model before $\rightarrow$ is the attack model, and the model after $\rightarrow$ is the victim model.}
\label{fig3}
\end{figure}

To explore the influence of JPEG compression, we used BIM, JPEGSS and BIM+IAM to create uncompressed adversarial examples on the LFW dataset and obtain the corresponding JPEG compressed adversarial examples with QF ranging from 10 to 90. 
And we followed the implementation of JPEGSS~\cite{JPEGSS} to generate adversarial examples with the preset QF (25). Fig.\ref{fig3} demonstrates the comparison of black-box performance between BIM, JPEGSS and our proposed interpolation attack method. Fig.\ref{fig3} indicates that lower QF values lead to more aggressive compression, which can significantly distort the adversarial examples and reduce the ASRs of them. As shown in Fig.\ref{fig3}, our proposed IAM maintains higher ASRs than BIM across different QF, and exhibits superior JPEG-resistance compared to JPEGSS in most cases.
\begin{figure}[!t]
\centerline{\includegraphics[width=8.65cm]{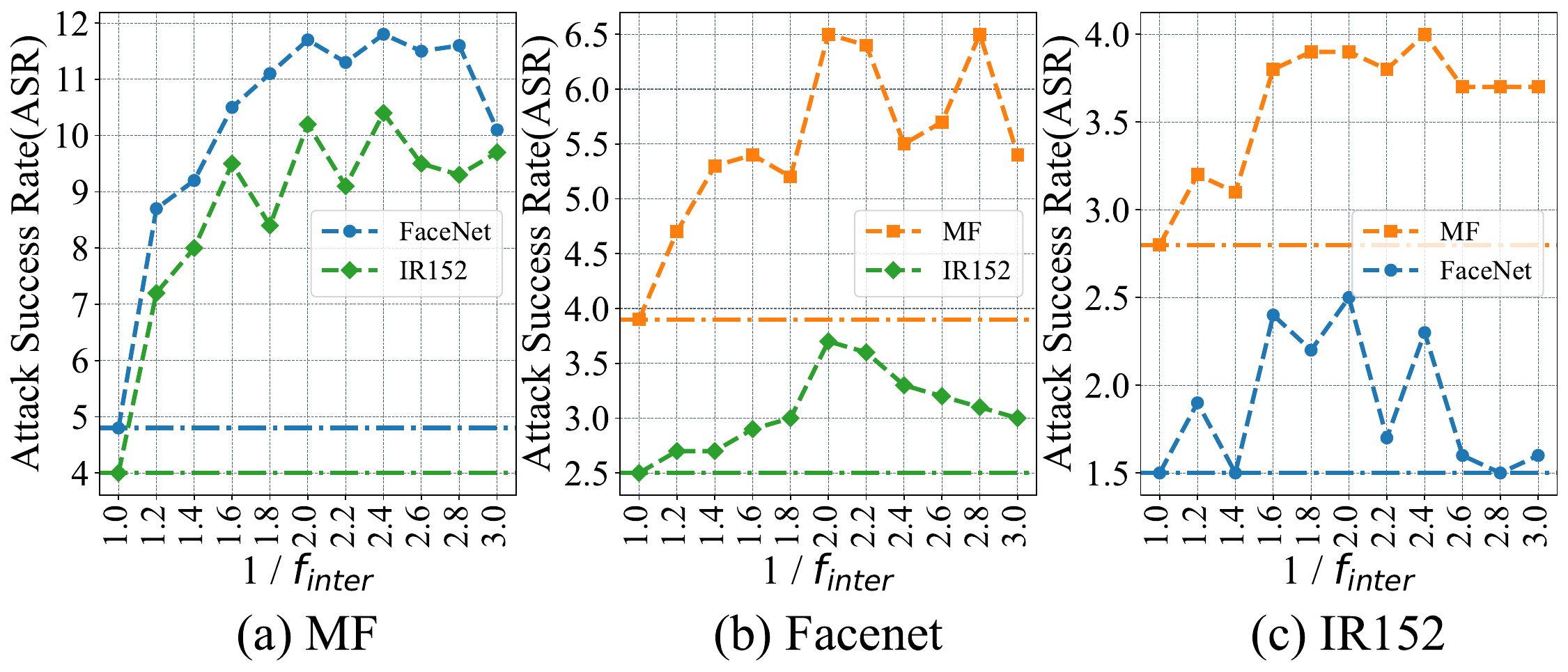}}
\caption{ASRs of the JPEG adversarial examples generated by BIM+IAM with different interpolation factors. In each subfigure, the subtitle is the corresponding attack model, the models in the legend are the corresponding victim models and the horizontal lines respresent the ASRs of BIM in the corresponding attack scenario.}
\label{fig4}
\end{figure}

Furthermore, we observed that JPEGSS exhibits satisfactory performance only when the QF is close to the preset QF, and the ASRs decrease as the QF continually increase. In Fig.\ref{fig3}, when employing IR152 as attacker models, the ASRs of JPEGSS are even lower than those of BIM when QF exceeds 40. This phenomenon occurs because the adversarial examples are specifically tailored to fit the preset QF of the differential JPEG approximation during the generation of JPEGSS, resulting in overfitting. This phenomenon contradicts the intended goal of improving the JPEG-resistance of adversarial examples. The results suggest that IAM can improve the resistance of adversarial attacks against JPEG compression across different compression levels, and the the performance of IAM is more comprehensive and stable compared to JPEGSS.
\subsection{Hyperparameter Sensitivity Study}
We adopted the conventional values for maximum iterations $N_{max}$, step size $\beta$, and perturbation magnitude $\epsilon$, as commonly used in prior studies. For the interpolation factor $f_{inter}$, we conducted a thorough experiment to investigate its impact on the ASRs of adversarial examples in black-box scenarios. All the generated adversarial examples undergo JPEG compression with QF of 50.

As Fig.\ref{fig4} shown, the ASRs of adversarial examples exhibit an increasing trend as the $\frac{1}{f_{inter}}$ rises, reaching a peak when the interpolation factor $f_{inter}$ is set to $\frac{1}{2}$.  When the interpolation factor $f_{inter}=1$, no interpolation operation is performed, and BIM+IAM degrades to BIM. When the interpolation factor $f_{inter}=\frac{1}{2}$, the ASRs of adversarial examples generated by BIM+IAM exhibit a significant improvement compared to BIM in all the attack scenarios. Addtionally, it is important to note that continuously increasing $\frac{1}{f_{inter}}$ beyond $2.0$ leads to unstable ASRs and a meaningless increase in computational complexity. Therefore, we set $f_{inter}=\frac{1}{2}$ to conduct more experiments.
\section{CONCLUSION}
In this paper, we present a novel adversarial attacks for improving the JPEG-resistance of adversarial attacks on FR. By utilizing interpolation smoothing, our proposed method can effectively decrease the low-frequency signals of the crafted adversarial face examples, thereby improving the JPEG-resistance of adversarial attacks on FR. Extensive experiments demonstrate the effectiveness of our proposed attack method.

\bibliographystyle{IEEEbib}
\bibliography{refs}

\end{document}